\documentclass{article}

\usepackage[nonatbib]{neurips_2022}

\usepackage[utf8]{inputenc} 
\usepackage[T1]{fontenc}    
\usepackage{hyperref}       
\usepackage{url}            
\usepackage{booktabs}       
\usepackage{amsfonts}       
\usepackage{nicefrac}       
\usepackage{microtype}      
\usepackage{xcolor}         
\usepackage{graphicx}
\usepackage{biblatex}
\usepackage{bbding}
\usepackage[lofdepth,lotdepth]{subfig}

\newcommand\blfootnote[1]{%
  \begingroup
  \renewcommand\thefootnote{}\footnote{#1}%
  \addtocounter{footnote}{-1}%
  \endgroup
}

\title{One Eye is All You Need: Lightweight Ensembles for Gaze Estimation with Single Encoders}

\author{%
  Rishi Athavale \\
  Vytal.ai \\
  McLean, VA \\
  \texttt{rishi.athavale1@gmail.com} \\
  \And
  Lakshmi Sritan Motati \\
  Vytal.ai, Harvard Medical School\\
  McLean, VA \\
  \texttt{lmotati@meei.harvard.edu} \\
  \And
  Rohan Kalahasty \Envelope \hspace{1pt} $^{ \dagger}$\\ 
  Vytal.ai, Harvard Medical School \\
  McLean, VA \\
  \texttt{rkalahasty@meei.harvard.edu} \\
}

\begin{document}

\maketitle
\blfootnote{$\dagger$ denotes corresponding author.}

\begin{abstract}
    Gaze estimation has grown rapidly in accuracy in recent years. However, these models often fail to take advantage of different computer vision (CV) algorithms and techniques (such as small ResNet and Inception networks and ensemble models) that have been shown to improve results for other CV problems. Additionally, most current gaze estimation models require the use of either both eyes or an entire face, whereas real-world data may not always have both eyes in high resolution. Thus, we propose a gaze estimation model that implements the ResNet and Inception model architectures and makes predictions using only one eye image. Furthermore, we propose an ensemble calibration network that uses the predictions from several individual architectures for subject-specific predictions. With the use of lightweight architectures, we achieve high performance on the GazeCapture dataset with very low model parameter counts. When using two eyes as input, we achieve a prediction error of 1.591 cm on the test set without calibration and 1.439 cm with an ensemble calibration model. With just one eye as input, we still achieve an average prediction error of 2.312 cm on the test set without calibration and 1.951 cm with an ensemble calibration model. We also notice significantly lower errors on the right eye images in the test set, which could be important in the design of future gaze estimation-based tools.
\end{abstract}

\section{Introduction}
In the past years, advances in gaze tracking technology have allowed for its use in a large variety of fields. An abundance of research has shown that many neurological diseases, including Parkinson's, Alzheimer's, Schizophrenia, and Autism, have distinctive, early onset bio-markers that can be found through eye movement analysis \cite{mao_he_liu_chen_2020}. For example, studies have shown that people with Alzheimer's have higher correction rates during rapid eye movements and fixations \cite{opwonya_doan_kim_kim_ku_kim_park_kim_2021}. Gaze tracking headsets can be used to diagnose diseases based on these bio-markers by analyzing saccadic changes. Gaze tracking is also heavily utilized for advertising research, where it is used to see which parts of the ad are focused on by users. However, the current state-of-the-art gaze tracking devices cost thousands of dollars, preventing the widespread adoption of the technology. Lower-cost gaze tracking methods need to be developed.

Current methods for gaze tracking leverage modern advancements in machine learning (ML), computer vision (CV), and graphics processing units (GPUs) to train ML models to make gaze predictions. While it is common for ML-based gaze estimators to mainly make use of frontal images of a subject, many methods have utilized less conventional forms of data such as video \cite{videogaze}, 3D models \cite{3dmodelgaze}, and 3D facial reconstructions \cite{3dfacereconstruction}.

More traditional methods of ML-based gaze estimation, however, instead use an image of a subject, with some using the full frontal face and some using cropped eye images. In order to improve performance, they often calibrate a model to a specific user by first collecting data from them and then tuning the model to the user-specific data \cite{mitpaper, googlepaper}. This allows models to fit to the specific features present in a specific user.

These methods, although somewhat simple, are generally less computationally intensive than methods that use more complex data inputs. Furthermore, they are also able to achieve impressive results by training on large amounts of data. One of the largest gaze prediction datasets is the GazeCapture dataset, which includes nearly 2.5 million frames taken from over 1450 people \cite{mitpaper}. In their paper "Eye Tracking for Everyone", Kafka et al. introduce not only the GazeCapture dataset but also an ML architecture to generate gaze estimations. Their method takes as input the face and eyes from a frame as well a face grid. The face image, right eye image, and left eye image are each passed through their own feed-forward convolutional neural network (CNN), which extracts their features. The features of the face grid are extracted using a feed-forward neural network (NN). The features for both of the eyes are then passed through a fully connected (FC) layer to generate features that represent both of the eyes. These feature vectors are then concatenated and passed through a feed-forward NN which outputs a \((x, y)\) coordinate that represents the gaze prediction. Without calibration, this method was able to achieve a prediction error of 1.71 cm on data from mobile phones and a prediction error of 2.53 on data from tablets \cite{mitpaper}.

Research from Valliappan et al. was able to achieve impressive results using the GazeCapture dataset while only using eye images and eye landmarks. They utilized a "dual tower" approach, in which each eye image (with the left eye being flipped horizontally) has its features extracted by the same CNN. They also used a feed-forward NN to extract features from eye landmarks, which represent the top left and bottom right coordinates of each eye. The eye features and landmark features are then concatenated together, after which they are passed through a feed-forward NN which outputs a \((x, y)\) coordinate that represents the gaze prediction. With this approach, Valliappan et al. were able to achieve a prediction error of 1.92 cm on the GazeCapture dataset \cite{googlepaper}.

Current gaze estimation methods generally have the advantage of being lightweight while being able to leverage large datasets to achieve low prediction errors. However, these methods fail to make use of other CV techniques that could both improve results while keeping gaze estimators lightweight. The ResNet and Inception architectures, for example, have achieved great success on many CV tasks such as the ImageNet challenge (on which ResNet achieved a top-5 error rate of 3.57\% in 2015 \cite{resnetpaper} and Inception achieved a top-5 error rate of 6.67\% in 2014 \cite{inceptionpaper}). Additionally, methods such as ensemble models (which combine predictions from multiple models) have helped improve results in CV tasks such as image classification \cite{ensemblepaper}. The fact that traditional gaze prediction methods don't utilize such methods could be limiting their performance.

Current gaze estimation methods also often rely on having an image of a person's face \cite{mitpaper} or an image of both a person's eyes \cite{googlepaper}. This limits their applicability because, in real-world conditions, it is not uncommon for a person's face or one of their eyes to be partially obscured, for an eye-detection model to only successfully detect one eye, or even for a user to be missing an eye. A more practical method of gaze estimation, therefore, would be one that can make a prediction from just a single eye.

\section{Methodology}
To test the effectiveness of other computer vision methods, we modified the CNN architecture in \cite{googlepaper} to make three new versions: one with a ResNet-based architecture, one with an Inception-based architecture, and one with an InceptionResNet-based architecture. In addition, we made modified versions of all of these models that would only take in one eye as input in order to test one-eye gaze estimation. After training and evaluating these models based on mean Euclidean distance (in centimeters), we calibrated them to testing set subjects by fitting a lightweight support vector machine (SVM) to training set frames for a specific subject.
\subsection{Data Preprocessing}
To train and evaluate models for gaze estimation, we used the GazeCapture dataset. This dataset includes nearly 2.5 million frames along with important metadata (such as the size of the screen and the location of the eyes and face) for over 1450 subjects \cite{mitpaper}. We use frames taken from iPhones in portrait orientation and with valid eye detections, resulting in a dataset of 501,675 frames. Similar to \cite{googlepaper}, we also modified the metadata for each frame to include automatically-extracted eye corner landmarks in order to provide the location of the subject's eyes in the frame. For each frame, the eye regions (with each one being resized to 128x128) and the eye landmarks (coordinates of the top left and bottom right corners of the eye regions) were used.

To split the dataset into training, validation, and testing sets, we used a similar method to \cite{googlepaper} in that we split our dataset based on unique ground truth points. This means that while frames from the same subject can appear in different datasets, frames corresponding to a particular ground truth point will not appear in different datasets (thereby preventing data leakage). Using an 80/8/12 dataset split, the training set had 399,093 frames, the validation set had 43,558 frames, and the testing set had 59,024 frames.

\subsection{Proposed Approach}
Current gaze estimation methods have largely opted for basic CNN architectures. However, more complex methods such as the ResNet and Inception architectures have been shown to perform better than traditional CNNs in a variety of tasks, such as the ImageNet challenge \cite{resnetpaper, inceptionpaper}. In order to integrate preexisting gaze estimation architectures with more complex CV methods, we propose modified versions of the model architecture used in \cite{googlepaper} in which the eye-feature-extractor (which in \cite{googlepaper} is a basic feed-forward CNN) is replaced with a small ResNet architecture, a small Inception architecture, or a small InceptionResNet architecture. By basing these models on the one used by \cite{googlepaper} and keeping the model architectures small, the proposed models continue to be lightweight while being able to better map eye images to features due to having more parameters.

A weakness of current gaze estimation methods is that they require either a frontal view of a subject's face or images of both of their eyes \cite{mitpaper, googlepaper}. This limits the practical use of these models because it makes them less useful in cases where a person's face is partly obscured or when their head is turned. A significantly more practical method would be one that could make a gaze estimation from an image of a single eye. Not only would this allow for more accurate gaze estimations in the aforementioned cases, but it would also be usable for frames in which a given eye-detection model can identify only one eye.

Current gaze calibration methods involve either the retraining of a full model or fitting a lightweight model to map the activations from the base model to the ground truth points of the subject \cite{googlepaper}. The first method is extremely impractical as not only does retraining a full model take significantly more time than a lightweight calibration model, but it would also be difficult to scale to a large user-base as each retrained model would have to be separately saved. While this could be avoided by re-calibrating the model with every use, this would be inconvenient and time-consuming for users. The second method is also limited by the fact that the lightweight calibration model can only base its predictions on features from the base model. This means that if there is a problem with the base model's features, it can be nearly impossible for the calibration model to make accurate predictions for a subject. To improve on the limitations of current calibration methods, we propose an ensemble approach in which the lightweight calibration model is fit on features from multiple base models. Ensemble models have proven to be successful on a variety of CV tasks (such as the ImageNet challenge \cite{ensemblepaper}) because of their ability to utilize information from multiple models. By using an ensemble approach, the lightweight calibration model will be less likely to be limited by problems with base model features because if multiple base models are used, it is more likely one of them will have features that can better fit the subject.
\subsection{Model Architectures}
The proposed model architectures are modified versions of the one used in \cite{googlepaper}. While the landmark feature extractor (the feed-forward neural network that processes the eye corner landmarks) is largely kept the same, the eye image feature extractor (the CNN that processes the eye images) is modified.
\subsubsection{Regular CNN architecture}
The regular CNN architecture (which in results will be referred to as "CNN") is based on the one presented in \cite{googlepaper}. The architecture consists of two CNN "towers" (which process the eye images) and a landmark model which processes the eye landmarks with a feed-forward neural network. One CNN tower processes the right eye image and the other processes the left eye image (which is flipped horizontally). The two CNN towers are the same model and share weights. Each tower consists of three convolutional layers with kernel sizes of 7x7, 5x5, and 3x3, output channels of 32, 64, and 128, and strides of 2, 2, and 1. Each convolutional layer is followed by an average pooling layer with a pooling size of 2x2. The output is of size 2x2x128, which when flattened is of size 512 \cite{googlepaper}.

The landmark model is a feed-forward neural network with three layers with 128, 16, and 16 units respectively. Each of these layers is followed by  batch normalization and a rectified linear unit (ReLU) layer \cite{googlepaper}.

\begin{figure}[h]
\centering
\subfloat[][CNN architecture that we test \cite{googlepaper}.]{
\includegraphics[width=0.8\textwidth]{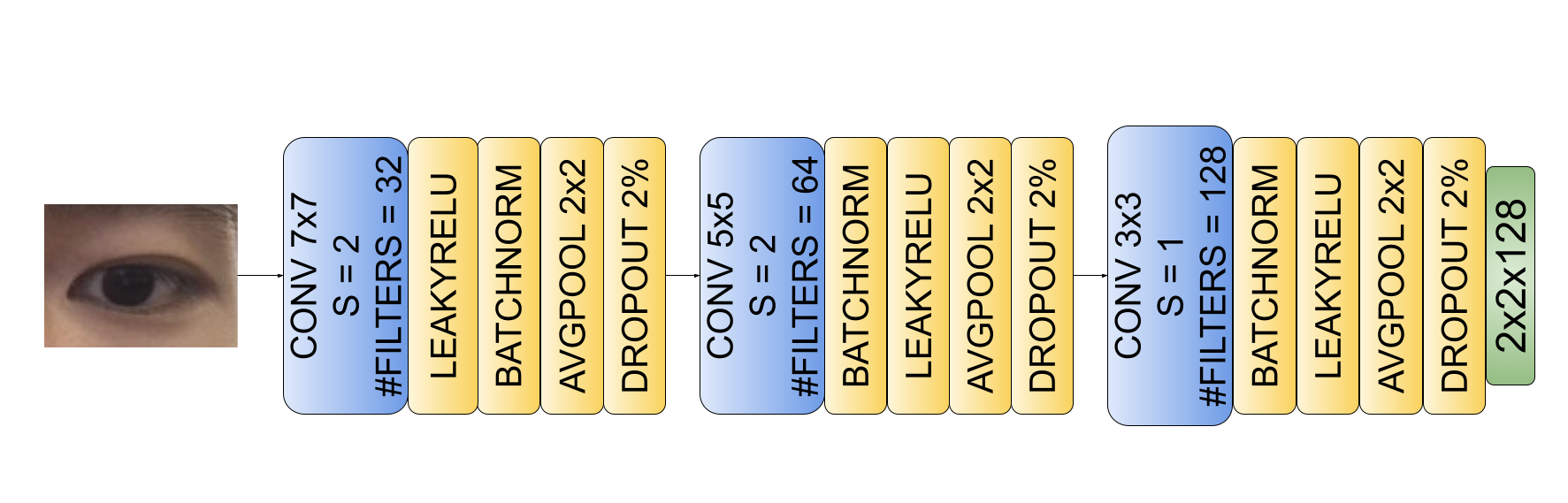}
\label{fig:googlemodel}}
\qquad
\subfloat[][ResNet architecture that we test.]{
\includegraphics[width=0.8\textwidth]{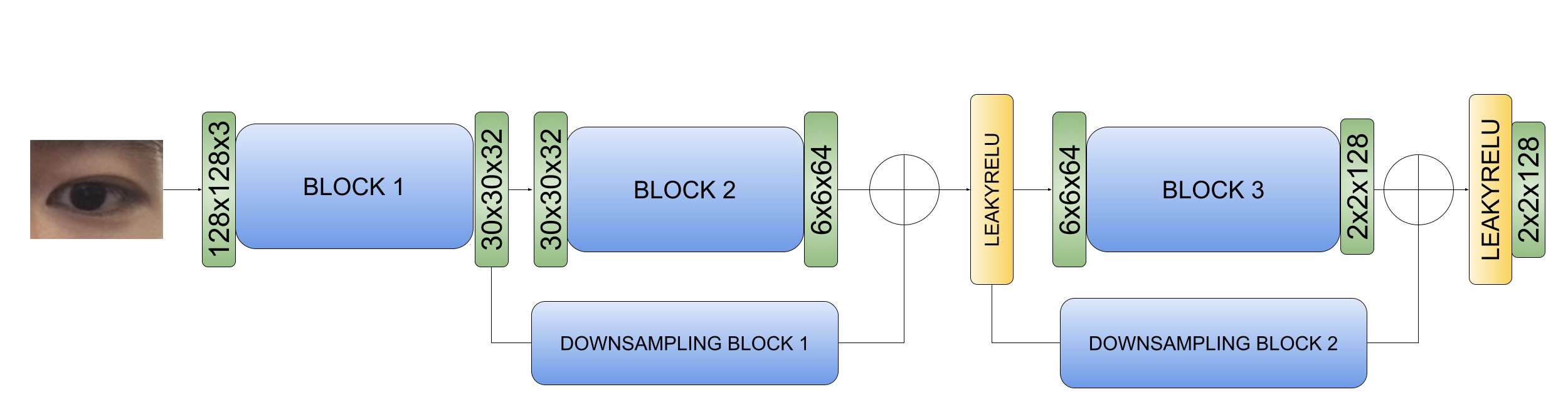}
\label{fig:resnet}}
\qquad
\subfloat[][Inception architecture that we test.]{
\includegraphics[width=0.8\textwidth]{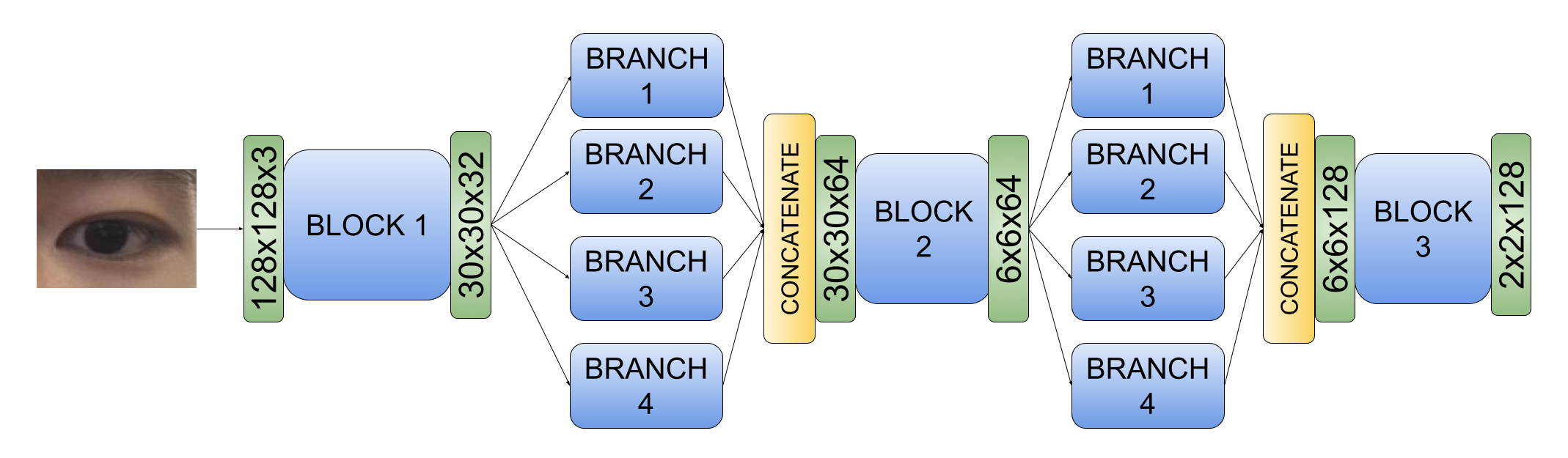}
\label{fig:inception}}
\qquad
\subfloat[][InceptionResNet architecture that we test.]{
\includegraphics[width=0.8\textwidth]{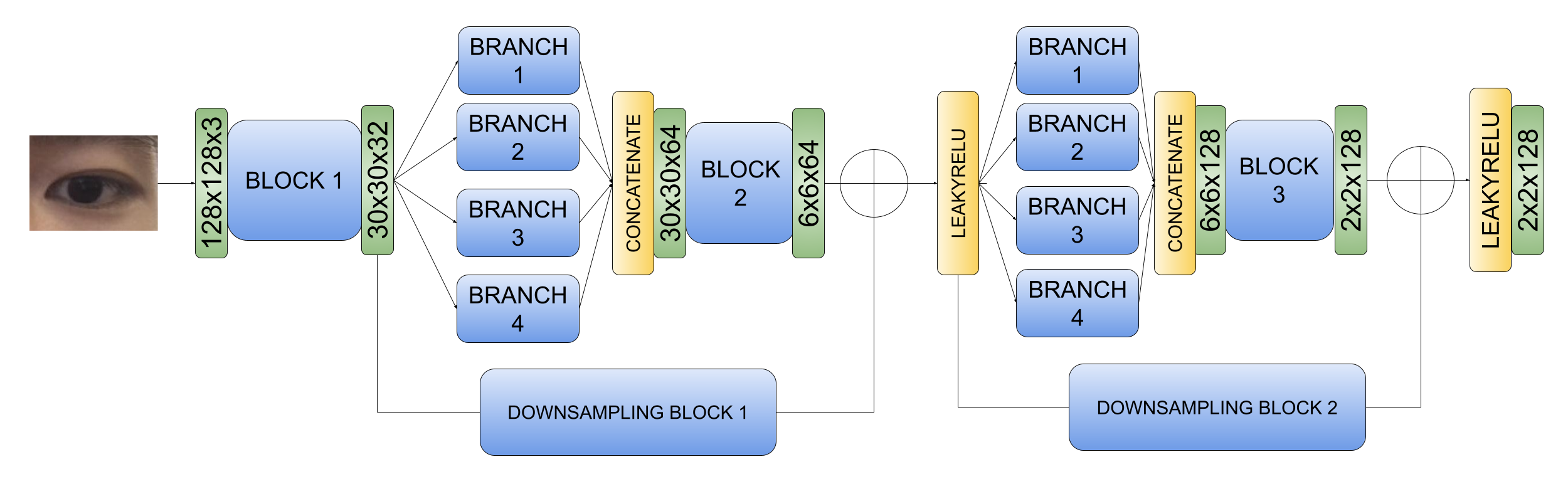}
\label{fig:inceptionresnet}}
\caption{Model architectures that we implement.}
\label{fig:architectures}
\end{figure}

The feature vectors from the eyes (each of size 512) are then concatenated with the feature vector from the eye landmarks (which has a size of 16). The resulting concatenated feature vector is then processed through a feed-forward neural network with three layers with 8, 4, and 2 units respectively \cite{googlepaper}. The CNN architecture used is shown in Figure \ref{fig:googlemodel}.


\subsubsection{ResNet architecture}
The ResNet architecture used includes only two residual blocks in order to keep the model small and lightweight. It consists of three blocks, with each consisting of two convolutional layers (with kernel sizes 4x4 for the first block, 3x3 for the second block, and 2x2 for the third block), two batch normalization layers (one after each convolutional layer), a leaky rectified linear unit (ReLU) layer (after the first convolutional layer), an average pooling layer with a pooling size of 2x2 (after the second convolutional layer) and a dropout layer (after the second convolutional layer). The second convolutional layer in each block had a stride of 2. The output channels for each block were 32, 64, and 128 respectively. In order to create residual blocks, the first and second blocks are added as are the second and third blocks. To do this, a down-sampling block is applied to the first block so it will be the same size as the second block, and another down-sampling block is applied to the second block so it will be the same size as the third block. The first down-sampling block has 64 output channels and consists of a convolutional layer with a kernel size of 5x5 and a stride of 2, a batch normalization layer, an average pooling layer with a pooling size of 2x2, and a dropout layer. The second down-sampling block is the same except it has 128 output channels and its convolutional layer has a kernel size of 3x3 and a stride of 1. The resulting output is of size 2x2x128 (which when flattened is of size 512). The diagram for the ResNet architecture used is shown in Figure \ref{fig:resnet}.


\subsubsection{Inception architecture}
The Inception architecture, like the ResNet architecture, consists of three blocks, two of which are inception blocks. The first block consists of a convolutional layer with a kernel size of 7x7, a stride of 2, and 32 output channels, a batch normalization layer, a leaky ReLU layer, an average pooling layer with a pooling size of 2x2, and a dropout layer. The second block consists of four branches that are concatenated together: one with one convolutional layer with a kernel size of 1x1, a stride of 1, and 32 output channels and a batch normalization layer, one with two convolutional layers (with kernel sizes of 1x1 and 3x3 respectively and each having a stride of 1 and 32 output channels) and a batch normalization layer, one with two convolutional layers (with the first one having a kernel size of 1x1 and the second having a kernel size of 5x5 and a padding value of 1, and each having a stride of 1 and 8 output channels) and a batch normalization layer, and one with a max pooling layer with a pooling size of 3x3, a stride of 1, and padding value of 1, a convolutional layer with a kernel size of 1x1, a stride of 1, and 8 output channels, and a batch normalization layer. After the four branches are concatenated, they are passed to a convolutional layer with a kernel size of 5x5, a stride of 2, and 64 output channels, a batch normalization layer, a leaky ReLU layer, an average pooling layer with a pooling size of 2x2, and a dropout layer. The four branches, concatenation, and subsequent layers are part of the second block. The outputs from the second block are then passed to the third block, which consists of four branches. These branches are the same as the ones in the second block, except that the number of output channels is doubled. The four branches, after being concatenated, are passed to a convolutional layer with a kernel size of 3x3, a stride of 1, and 128 output channels, a batch normalization layer, a leaky ReLU layer, an average pooling layer with a pooling size of 2x2, and a dropout layer. The resulting output is of size 2x2x128 (which when flattened is of size 512). The diagram for the Inception architecture used is shown in Figure \ref{fig:inception}.


\subsubsection{InceptionResNet architecture}
The InceptionResNet architecture integrates the aforementioned ResNet and Inception architectures. Its architecture is the same as the Inception architecture described above, but with residuals that are obtained by down-sampling the first and second blocks and added to the second and third blocks respectively. Since the shapes of the outputs of each block are the same, the down-sampling blocks are the same as the ones described in the ResNet architecture. The diagram for the InceptionResNet architecture used is shown in Figure \ref{fig:inceptionresnet}.

\subsubsection{One-Eye Gaze Estimation}
The one-eye gaze estimation models use the same eye image feature extraction architectures (the CNNs that process the eye images) as those used in the two-eye gaze estimation models. However, they will only process one eye image (as opposed to the two-eye gaze models, which use a "dual tower" approach to process both eye images). When left-eye images are passed to the one-eye gaze estimators, they are not flipped horizontally. This is because flipping the left eye image horizontally when only one eye image is given would make it harder for the model to determine what direction the eye is looking in. For example, if a left eye that is looking to the left is given to the model and it is flipped horizontally, it would be difficult for the model to tell if it was a left eye looking leftward or a right eye looking rightward. The landmark feature extractor (the feed-forward neural network that processes the eye-corner landmarks) is largely the same, except that the input shape is of size 4 due to only processing 2 coordinates.

During training, each frame is randomly assigned a 0 or a 1. For frames that are assigned a 0, their right eye image and right eye landmarks will be used, and for frames that are assigned a 1, their left eye image and left eye landmarks will be used.
\subsection{Ensemble-based Calibration}
Since data was split based on ground truth point rather than subject, each subject has a number of frames in the training and the test set. For each frame of a subject present in the training set, each model's (regular CNN, ResNet, Inception, and InceptionResNet) activations before the linear output layer are collected and concatenated into a feature vector. This results in a set of feature vectors that each correspond to a ground truth value. A support vector machine (SVM) is then trained to map the feature vectors to their respective ground truth point. After the SVM is trained, the feature vectors are then obtained for the subject's frames present in the test set. The SVM's predictions are then compared with the ground truth points using the mean Euclidean distance in centimeters. This process is done for every subject in the dataset to determine the average performance of the calibration method across the dataset.

\begin{figure}
  \includegraphics[width=\linewidth]{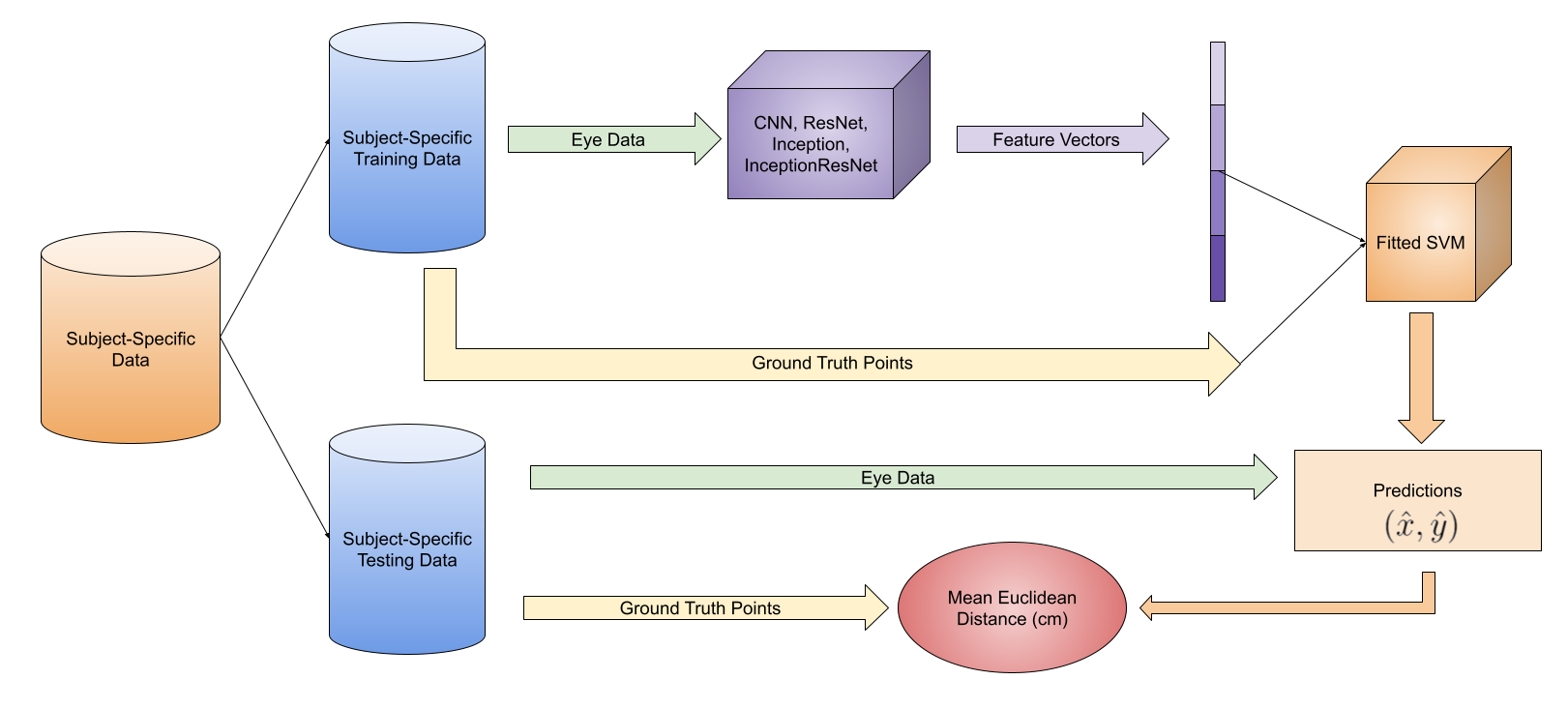}
  \caption{Diagram for ensemble-based calibration.}
  \label{fig:calibration}
\end{figure}

For the one eye models, calibration is tested in three different ways. First, the process described above is performed but only with data for the subjects' right eyes (meaning that feature vectors are obtained by passing the right eye image and right eye landmarks corresponding to a frame). This is then done again but with data for the subjects' left eyes. For the final method, both eyes are used. This is done by getting the feature vectors for the right eye data and the feature vectors for the left eye data and concatenating them together. This concatenated feature vector is then used as the input to the SVM, which is trained to map the concatenated feature vectors to the ground truth points. The diagram for the ensemble-based calibration is shown in Figure \ref{fig:calibration}.

\subsection{Training}
The models were all trained with the Adam optimizer, an initial learning rate of 0.016, an exponential learning rate scheduler, and a batch size of 256. The training loss was the mean squared error between the predicted coordinates and the true coordinates. Mean squared error was also used to evaluate the model's performance on the validation set (with a lower mean squared error indicating a better model and a higher mean squared error indicating a worse model). Training was done for 50 epochs, and the weights with which it performed best on the validation set (obtained the lowest validation mean squared error) were saved.

\subsection{Evaluation}
To evaluate the quality of a gaze prediction, the Euclidean distance (in centimeters) between the predicted coordinate \((\hat{x}, \hat{y})\) and the true coordinate \((x, y)\) is obtained, and it is calculated using the following formula: \[\sqrt{(\hat{x} - x)^2 + (\hat{y} - y)^2}\]
To evaluate a gaze estimator's performance on a dataset, the mean of the Euclidean distances for every frame is calculated.

\subsection{Implementation Details}
\label{section:implementation}
The developed models were trained and evaluated on the GazeCapture dataset, which was split into train, validation, and test sets. The models were made using the PyTorch framework. The training was done with an AWS p3.2xlarge instance, which included a Tesla V100 GPU with 16 GB of memory. Our code can be found here: \href{https://github.com/rishipython/One-Eye-is-All-You-Need-Lightweight-Ensembles-for-Gaze-Estimation-with-Single-Encoders}{GitHub}.

\section{Experiments and Results}
\begin{table}
  \caption{Mean Euclidean distance (in centimeters) for each model with two eyes as input on the validation and testing sets.}
  \centering
  \begin{tabular}{lll}
    \toprule
    Model     & Validation Mean Error (cm)     & Test Mean Error (cm) \\
    \midrule
    CNN & 1.694  & 1.692     \\
    ResNet     & 1.604 & 1.607      \\
    Inception     & 
    \textbf{1.590}       & 1.591  \\
    InceptionResNet     & 1.683       & 1.683  \\
    Calibration & N/A & \textbf{1.439} \\
    \bottomrule
  \end{tabular}
  \label{table:table1}
\end{table}

\begin{table}
  \caption{Mean Euclidean distance (in centimeters) for each model with one eye as input on the validation and testing sets.}
  \centering
  \begin{tabular}{llll}
    \toprule
    Model     & Eye Data     & Validation Mean Error (cm)   & Test Mean Error (cm) \\
    \midrule
    CNN & Right Eye & 2.073 & 2.078 \\
    ResNet & Right Eye & 2.000 & 2.013 \\
    Inception & Right Eye & \textbf{1.934} & \textbf{1.947} \\
    InceptionResNet & Right Eye & 2.042 & 2.049 \\
    \midrule
    CNN & Left Eye & 2.647 & 2.685 \\
    ResNet & Left Eye & \textbf{2.605} & \textbf{2.611} \\
    Inception & Left Eye & 2.678 & 2.708 \\
    InceptionResNet & Left Eye & 2.651 & 2.691 \\
    \midrule
    Calibration & Right Eye & N/A & \textbf{1.800} \\
    Calibration & Left Eye & N/A & \textbf{2.102} \\
    Calibration & Both Right and Left Eye & N/A & \textbf{1.774} \\
    
    \bottomrule
  \end{tabular}
  \label{table:table2}
\end{table}

\begin{table}
  \caption{Average of left-eye and right-eye mean Euclidean distances (in centimeters) for the one eye models on the validation and testing sets.}
  \centering
  \begin{tabular}{lll}
    \toprule
    Model     & Average Validation Mean Error (cm)   & Average Test Mean Error (cm) \\
    \midrule
    CNN & 2.360 & 2.382 \\
    ResNet & \textbf{2.303} & 2.312 \\
    Inception & 2.306 & 2.327 \\
    InceptionResNet & 2.346 & 2.370 \\
    Calibration & N/A & \textbf{1.951} \\
    
    \bottomrule
  \end{tabular}
  \label{table:table3}
\end{table}

    

\begin{figure}[h]
\centering
\subfloat[][Training loss of two eye models.]{
\includegraphics[width=0.45\textwidth]{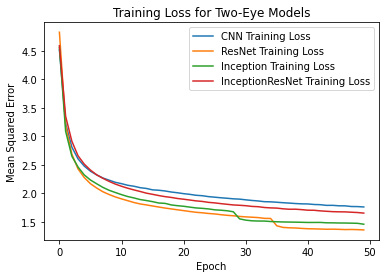}
\qquad
\label{fig:subfig1}}
\subfloat[][Validation loss of two eye models.]{
\includegraphics[width=0.45\textwidth]{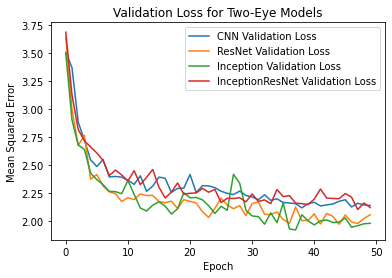}
\label{fig:subfig2}}
\qquad
\subfloat[][Training loss of one eye models.]{
\includegraphics[width=0.45\textwidth]{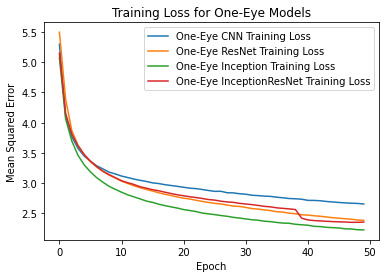}
\label{fig:subfig3}}
\subfloat[][Validation loss of one eye models.]{
\includegraphics[width=0.45\textwidth]{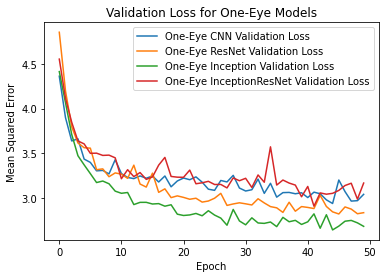}
\label{fig:subfig4}}
\caption{Graphs of the training or validation loss over training for the two eye and one eye models.}
\label{fig:graphs}
\end{figure}

The four different models (CNN, ResNet, Inception, and InceptionResNet) were trained on the training set for 50 epochs whilst saving the model weights that achieved the lowest validation loss. This was done for both the two-eye and one-eye models. Figure \ref{fig:graphs} shows the training and validation curves for each model. As shown in Table \ref{table:table1}, the Inception model did the best on the validation and test sets out of the four base models, with it achieving a validation mean error of 1.590 cm and a test mean error of 1.591 cm. Furthermore, the regular CNN model did the worst out of the four. The low variability between the validation and the test mean errors show that the models likely did not overfit.

As shown in Table \ref{table:table2}, the Inception model achieved the lowest errors on the validation and test sets when given only right eye data (1.934 cm on the validation set and 1.947 cm on the test set) and the ResNet model achieved the lowest errors on the validation and test sets when given only left eye data (2.605 cm on the validation set and 2.611 cm on the test set). One-eye models tended to perform significantly better when given right eye data as opposed to left eye data. As shown in \ref{table:table3}, the ResNet model performs the best on average out of the four base models, with its average error being 2.303 cm on the validation set and 2.312 cm on the test set.

For both the two-eye and one-eye models, ensemble calibration with a lightweight SVM was able to dramatically reduce error. As shown in Table \ref{table:table1}, the two-eye calibration model achieved a test mean error of 1.439 cm (0.152 cm less than Inception's test mean error). The one-eye calibration model, as shown in Table \ref{table:table2}, achieved a right eye mean error of 1.800 cm and a left eye mean error of 2.102 cm, the average of which (as shown in Table \ref{table:table3}) is 1.951 cm (0.361 cm less than the average error for ResNet on the test set). Furthermore, the one-eye calibration model was able to achieve a test mean error of 1.774 cm when the SVM was given concatenated feature vectors from both left and right eye data, as shown in Table \ref{table:table2}.

\section{Conclusion}

The results of this study show the effectiveness of the ResNet and Inception architectures as well as ensemble calibration for gaze tracking. We were able to show that ResNet, Inception, and InceptionResNet models were able to significantly outperform standard CNN architectures, which are commonly used for gaze estimation. Our Inception base model, which achieved a test mean error of 1.591 cm, was also able to outperform the base models created by \cite{mitpaper} (which achieved an error of 1.71 cm) and \cite{googlepaper} (which achieved an error of 1.92 cm) for GazeCapture. We were also able to show how small versions of the ResNet, Inception, and InceptionResNet architectures could make accurate gaze predictions while remaining lightweight.

Additionally, we were able to show how gaze estimators can still be fairly accurate even when only given data for a single eye. Despite the one-eye gaze models having half as much data per example as the two-eye gaze models, they were able to achieve an average error of 2.312 cm without calibration (based on ResNet performance) and 1.951 cm with calibration. When given data for both eyes, the calibration model performs even better, achieving an error of 1.774 cm. Not only does this show the high accuracy of one-eye gaze estimation, but it also shows that the one-eye gaze models were able to decipher unique and useful features for left and right eyes (as otherwise, the one-eye calibration model would not have seen a significant decrease in test mean error when using data from both eyes).

Further research should seek to implement the calibration methodology outlined in this paper with real-world video data. This would be done by recruiting participants to record 30-60 seconds on video where their gaze coordinates will be tracked (likely by instructing them to look at a specific point). This data can then be used to calibrate the calibration model. This would likely achieve better calibration results than the ones obtained in this paper, as we were limited to using data from subjects already in the GazeCapture dataset. Having video data where a participant is instructed to look at a specific point can be used to get a wider variety of frames and ground truth points for a participant. It would also be more accurate to how gaze calibration would be implemented in the real world. Other research should also seek to test other CV model architectures such as EfficientNet and MobileNet. Lightweight versions of these models would have to be created (similar to how lightweight versions of ResNet, Inception, and InceptionResNet were created for this study). Additionally, future research should seek to pretrain the CNN models on the ImageNet database and then retrain them on GazeCapture in order to see if transfer learning would improve results by allowing the models to start with an understanding of a variety of features before training. Pretraining models on ImageNet has been used for other gaze tracking models, such as L2CS-Net, to achieve impressive results \cite{l2csnet}.

The results of this study have important implications for gaze tracking with machine learning. They not only show that future research should focus on more complex CNN architectures such as ResNet and Inception instead of just regular CNNs, but they also show how one-eye gaze tracking can be a viable method for gaze prediction. This would allow impaired individuals with only a single eye to use gaze tracking software. It also has the potential to make gaze tracking software more applicable to real-world scenarios. In real-world data, it is common to have images where people aren't looking straight at a camera, one of their eyes is blocked by long hair, the eye detection software only detects a single eye for a given frame, and more cases where gaze predictions would have to be made based on a single eye. The findings of this research and the models produced have the potential to make gaze tracking more applicable to e-commerce, digital accessibility, and medical diagnosis.


\printbibliography

\end{document}